\documentclass{article} 
\usepackage[preprint]{colm2026_conference}

\usepackage{microtype}
\usepackage{hyperref}
\usepackage{url}
\usepackage{booktabs}
\usepackage{amsmath}
\usepackage{pdfpages}
\usepackage{wrapfig}
\usepackage{booktabs} 
\usepackage{siunitx}
\usepackage{lineno}
\usepackage{xcolor}
\usepackage[most]{tcolorbox}
\usepackage{float}
\usepackage{arydshln}
\usepackage{multirow}
\usepackage{wrapfig}

\definecolor{darkblue}{rgb}{0, 0, 0.5}
\hypersetup{colorlinks=true, citecolor=darkblue, linkcolor=darkblue, urlcolor=darkblue}

\newcommand{\modelname}{\texttt{DeIllusionLLM}} 
\newcommand{\dataname}{\texttt{FaultyScience}}   

\title{Bridging the Know–Act Gap via Task-Level Autoregressive Reasoning}

\author{
\textbf{Jihyun Janice Ahn}$^{1}$, \quad
\textbf{Ryo Kamoi}$^{1}$, \quad
\textbf{Berk Atil}$^{1}$, \quad
\textbf{Renze Lou}$^{1}$, \quad
\textbf{WonWoo Kang}$^{2}$ \\
\textbf{Heehyun Park}$^{1}$, \quad
\textbf{Sarkar Snigdha Sarathi Das}$^{1}$, \quad
\textbf{Zhuoyang Zou}$^{1}$, \quad
\textbf{Xiaoxin Lu}$^{1}$ \\
\textbf{Yusen Zhang}$^{3}$, \quad
\textbf{Asfahan Shah}$^{1}$, \quad
\textbf{Ridwanul Hasan Tanvir}$^{1}$, \quad
\textbf{Lingxiao Zhao}$^{1}$ \\
\textbf{Hongxi Huang}$^{1}$, \quad
\textbf{Vignesh Venkatesh}$^{1}$, \quad
\textbf{Dianjun Lin}$^{1}$, \quad
\textbf{Hamid Shah}$^{1}$ \\
\textbf{Wentao Wang}$^{1}$, \quad
\textbf{Zhanpeng Song}$^{1}$, \quad
\textbf{Joshua Reed Bassin}$^{1}$, \quad
\textbf{Dax Patel}$^{1}$ \quad \\
\textbf{Ishan Appareddy Agrahar}$^{1}$, 
\textbf{Sahil Pardasani}$^{1}$, \quad
\textbf{Xin Dong}$^{1}$, \quad
\textbf{Fatemeh Rahbari}$^{1}$ \quad \\
\textbf{Benjamin David Rishel}$^{1}$, 
\textbf{Soochan Andrew Lee}$^{1}$, \quad
\textbf{Yuv Boghani}$^{1}$, \quad
\textbf{Ali B. AlNaseeb}$^{1}$ \quad \\
\textbf{Pranav Suby}$^{1}$, 
\textbf{Seokhyeon Bae}$^{1}$, \quad
\textbf{Shreya Buddharaju}$^{1}$, 
\textbf{Damien Kula}$^{1}$, \\
\textbf{Soumyadeep Das}$^{1}$, \quad
\textbf{Hanyang Frank Liu}$^{3}$\thanks{Worked on this project when he was a student at Conestoga High School.}, \quad
\textbf{Faye Mo}$^{4}$, \quad
\textbf{Wenpeng Yin}$^{1}$ \\
\\
$^{1}$Penn State \quad
$^{2}$UIUC \quad
$^{3}$Columbia University \quad
$^{4}$Episcopal Academy \\
\texttt{\{jfa5672,wenpeng\}@psu.edu}
}

\begin{document}

\ifcolmsubmission
\linenumbers
\fi

\maketitle

\begin{abstract}
LLMs often generate seemingly valid answers to flawed or ill-posed inputs. This is not due to missing knowledge: under \emph{discriminative} prompting, the same models can mostly identify such issues, yet fail to reflect this in standard \emph{generative} responses. This reveals a fundamental \textbf{know--act gap} between discriminative recognition and generative behavior. Prior work largely characterizes this issue in narrow settings, such as math word problems or question answering, with limited focus on how to integrate these two modes.
In this work, we present a comprehensive analysis using \dataname{}, a newly constructed large-scale, cross-disciplinary benchmark of faulty scientific questions. We show that the gap is pervasive and stems from token-level autoregression, which entangles task selection (validate vs.\ answer) with content generation, preventing discriminative knowledge from being utilized. To address this, we propose \modelname{}, a task-level autoregressive framework that explicitly models this decision. Through self-distillation, the model unifies discriminative judgment and generative reasoning within a single backbone. Empirically, \modelname{} substantially reduces answer-despite-error failures under natural prompting while maintaining general reasoning performance, demonstrating that \textbf{self-distillation is an effective and scalable solution for bridging the discriminative--generative know--act gap}.\footnote{Data \&\ Code available: https://github.com/Janice-ahn/Validate-Before-You-Answer}
\end{abstract}

\section{Introduction}

The deployment environments of AI systems, particularly LLMs, are inherently unpredictable, involving varying conditions, constraints, and potential adversarial inputs. As AI becomes increasingly pervasive, ensuring its trustworthiness under such imperfect and conflicting scenarios is critical. A key question is how these systems behave when the input itself is flawed. Humans, for instance, typically validate a premise before proceeding, challenging inconsistencies or requesting missing information when necessary \citep{grice1975logic,johnson2010mental,byrne2019human,stalnaker2002common}. Do LLMs exhibit similar behavior?

Two empirical phenomena motivate this work. First, as illustrated in Table~\ref{tab:GPT_failure_example}, even top-performing LLMs often produce a direct answer under standard \textit{generative mode} without recognizing that the input itself is flawed. Second, when explicitly asked to judge whether the same input is problematic, i.e., under a \textit{discriminative mode}, the model can often correctly identify the issue. This contrast reveals a striking gap between \emph{what LLMs know} and \emph{what they actually do} during ordinary response generation. This observation raises two research questions ($\mathcal{Q}$):
\begin{itemize}
\item \textbf{$\mathcal{Q}_1$: How prevalent is the discrepancy between discriminative fault recognition and generative answering across mainstream LLMs, and why can models recognize an issue when explicitly asked, yet fail to leverage that knowledge under normal generation?}
\item \textbf{$\mathcal{Q}_2$: How can we incorporate LLMs' discriminative knowledge into the generative process so that the model can produce the most appropriate response \emph{without} human hints, explicit fault-handling instructions, or external policies?}
\end{itemize}

To answer $\mathcal{Q}_1$, we first introduce \dataname{}, a large-scale benchmark of over 15K faulty questions, primarily from science and engineering, spanning eight disciplines including physics, biology, earth science, mathematics, computer science, medicine, social science and more. The dataset covers diverse question formats (e.g., open-ended, multiple-choice) and error types (e.g., missing information, incorrect premises, nonsensical content). Using this benchmark, we systematically compare LLM behavior under three prompting settings: natural generative prompting without hints, generative prompting with a subtle warning, and explicit discriminative evaluation. We find that while models such as Llama \citep{touvron2023llamaopenefficientfoundation}, Qwen \citep{qwen2025qwen25technicalreport}, and GPT4 \citep{openai2024gpt4technicalreport} achieve around 90\% accuracy in discriminative fault recognition, their performance under the most natural no-hint generative setting drops dramatically---to slightly above 10\% for open-source models and only 34.9\% for GPT-4. Our analysis further reveals several recurring failure patterns: i) models often prioritize surface-level mathematical computation over basic validity checks, ii) answer only the seemingly solvable parts of a flawed question, or iii) default to producing a helpful-looking response instead of acknowledging that the problem itself is invalid.

\begin{table}
\centering
\small
\setlength{\tabcolsep}{5pt}
\renewcommand{\arraystretch}{1.25}

\begin{tabular}{p{0.25\linewidth}|p{0.7\linewidth}}
\hline

Question & 
Balanced binary data: accuracy 95\%, prec. 90\%, recall 85\%, compute F1.\\
\hline

Ask GPT4 to respond (w/o hint)&
\multirow{2}{*}{$F1=\frac{2PR}{P+R}=\frac{2\times 0.90\times 0.85}{0.90+0.85}=0.874$}\\
\hline

\multirow{3}{*}{\parbox{1.0\linewidth}{ Ask GPT4 if the question is problematic}} &
On a balanced dataset, these metrics (accuracy = 95\%, precision = 90\%, recall = 85\%) cannot all hold simultaneously. At least one is invalid, so any F1 derived is unreliable.\\
\hline
\end{tabular}
\vspace{-3mm}
\caption{Example where GPT4 fails in generative mode without hints but recognize the issues in discriminative mode.}
\label{tab:GPT_failure_example}
\vspace{-3mm}
\end{table}
These findings motivate our investigation of $\mathcal{Q}_2$. We suspect that this failure cannot be fully addressed within standard token-level autoregressive generation, because the missing behavior is fundamentally \emph{task-level}: the model should first validate the input, then decide whether to answer, object, or request clarification. Based on this view, we propose \modelname{}, a model that distills both discriminative and generative behaviors into a unified LLM through task-level autoregressive training. Distilling from a strong teacher such as GPT improves an open-source backbone such as Qwen, but more strikingly, we find that self-distillation from Qwen's own discriminative and generative outputs also substantially improves its behavior on \dataname{}, while largely preserving performance on standard reasoning benchmarks such as BBH \citep{suzgun-etal-2023-challenging} and GPQA \citep{rein2024gpqa}. This suggests that self-distillation across reasoning modes is a practical and effective way to mitigate unconditional answering on adversarial prompts.

Overall, this work makes three main contributions. First, we provide the first systematic large-scale study of this trustworthiness-critical failure mode through the \dataname{} benchmark and extensive empirical analysis. Second, we propose \modelname{}, a task-level autoregressive training framework that enables LLMs to reason at the level of \emph{which task to perform} before generating content. Third, we show that mode-aware self-distillation can substantially reduce this failure mode without sacrificing general reasoning capability.

\section{Related Work}

\paragraph{Unanswerable and ambiguous but valid questions.}
Prior work has extensively studied how AI systems handle questions that are valid in form but unanswerable due to missing evidence or ambiguity. For example, \citet{sun2018unetmachinereadingcomprehension} propose U-Net, a unified model that jointly predicts answer spans and whether a question is unanswerable. Benchmarks such as SQuAD 2.0 \citep{Rajpurkar2018KnowWY} and subsequent analyses \citep{sulem-etal-2021-know-dont} evaluate whether models can abstain when no answer is supported by the given context. Other work \citep{min-etal-2020-ambigqa,elgohary-etal-2019-unpack} addresses ambiguity, where questions admit multiple valid interpretations, requiring models to disambiguate or enumerate possible answers. 

A key distinction from our work is that these settings assume the question itself is \emph{well-posed}. The challenge lies in missing information or multiple interpretations, emphasizing answer correctness rather than the model’s ability to verify the validity of the question itself.

\paragraph{LLMs on questions with flawed premises.}
More recent studies investigate how LLMs behave when faced with questions containing incorrect or misleading premises. These works show that LLMs often generate plausible but incorrect answers instead of rejecting invalid inputs \citep{Hu2023WontGF}. Similarly, benchmarks involving unsolvable mathematical or logical problems demonstrate that models tend to fabricate solutions rather than identify ill-posedness \citep{Sun2024BenchmarkingHI,rahman2025blindsolverslogicalthinkers}. 

However, these studies primarily characterize failure under standard generative prompting. They do not examine whether models internally possess the knowledge to recognize such flaws (e.g., under a discriminative setting), nor how this knowledge can be effectively utilized during generation.

\paragraph{Discriminative detection of problematic inputs.}
Another line of work treats problem validity as a standalone discriminative task. For instance, \citet{Lavi2025DetectingI} show that answerability can be detected via linear directions in embedding space, suggesting that LLMs implicitly encode solvability signals. Other studies focus on ambiguity detection and uncertainty calibration, showing that models can identify underspecified queries and adjust confidence \citep{Shi2025AmbiguityDA}. Related work also uses LLM representations to assess semantic validity in structured evaluation settings \citep{Milano2025ComparingHE}.

While these approaches demonstrate that LLMs can detect problematic inputs, they treat validation as an external classification task. They do not integrate this capability into the model’s reasoning process during generation.

\paragraph{Handling jailbreak attacks.} The most related work is \citep{ding-etal-2025-act}, which studies the discriminative–generative gap in the context of jailbreak prompts. Their approach introduces \textit{explicit guidance signals} (e.g., detection outputs or injected states) to control generation, operating in a \textit{with-hint} regime. In contrast, we address a broader \textit{know–act gap} and enable models to act on their knowledge \textit{under natural prompting without external hints}. Methodologically, they condition on auxiliary signals, while we model \textit{task selection (validate vs.\ answer) as an intrinsic autoregressive decision}, unifying discrimination and generation. The paradigm in \citep{ding-etal-2025-sdgo}  is generate $\rightarrow$ critique $\rightarrow$ refine, which depends on externalized feedback signals, while ours is decide $\rightarrow$ act within a single autoregressive trajectory, eliminating the need for post-hoc correction.

\paragraph{Novelty of our work.}
Our work differs from prior research in three key aspects. First, rather than evaluating whether LLMs \emph{can} detect problematic questions under explicit prompts or external probes, we focus on a practical failure mode: models often fail to invoke this capability under standard generative prompting without hint. 
 Second, existing datasets typically focus on domain-specific cases, such as math word problems or jailbreak attacks. We introduce \textsc{FaultyScience}, a large-scale, cross-disciplinary benchmark of naturally occurring faulty scientific questions, designed to evaluate whether models can autonomously decide whether to answer or object under realistic no-hint conditions. Third, beyond analysis, we propose a solution. Our method, \textsc{DeIllusionLLM}, introduces task-level autoregressive reasoning and a scalable self-distillation paradigm that integrates discriminative validation and generative answering within a unified model.

\section{\dataname{} Dataset Construction}
\label{sec:dataset}

We construct \dataname{}, a large-scale benchmark of \emph{flawed} scientific questions that are plausible at first glance but ultimately ill-defined or unsolvable. The design goal is to evaluate whether LLMs can \emph{self-initiate} doubt and refuse, diagnose, or request clarification under realistic prompting, rather than succeeding only when explicitly warned that an input may be invalid.

\paragraph{Data Collection.} We curate candidate questions from a pool of $\sim$100 undergraduate \& graduate students spanning multiple scientific disciplines from some courses the author team taught. Contributors are instructed to provide \emph{science questions that contain natural errors}, e.g., incorrect premises, contradictory constraints, missing conditions, or scientifically impossible setups, rather than artificially constructed adversarial prompts. This collection strategy yields diverse error patterns that better reflect how ill-posed questions arise in real educational and scientific settings (e.g., misunderstanding of concepts, unit/quantity mismatches, ambiguous variable definitions, or invalid multiple-choice options).

\paragraph{Manual Filtering.}
 Raw submissions include both non-scientific items and trivial failures that can be detected by superficial cues. To ensure the benchmark emphasizes meaningful reasoning and validation, we apply a multi-stage manual filtering pipeline: i) \textit{Non-science removal.} We discard prompts that are not scientific or technical questions (e.g., general chit-chat, purely opinion-based prompts, or requests unrelated to scientific problem solving). ii) \textit{Trivial-fault removal.} We remove questions whose faultiness is obvious without substantive reasoning (e.g., blatant nonsense strings, malformed text, or errors detectable from a single shallow pattern). iii) \textit{Well-posedness screening.} We prioritize questions that appear solvable on the surface yet contain a subtle fatal flaw such as: (i) incorrect scientific premise, (ii) internal logical contradiction, (iii) missing essential conditions, (iv) invalid answer choices, or (v) impossible scenario under standard scientific knowledge.

\begin{wraptable}[23]{r}{0.48\textwidth}
\centering
\vspace{-4mm}
\setlength{\tabcolsep}{2pt}
\begin{tabular}{lrr}
\toprule
 & \textbf{Count} & \textbf{Ratio (\%)} \\
\midrule
\textbf{Discipline} & & \\
\addlinespace[0.5em]
Physical Science & 3014 & 19.07 \\
Biological Science & 2776 & 17.57 \\
Earth Science & 2404 & 15.21 \\
Mathematics & 2129 & 13.47 \\
Social Science & 2070 & 13.10 \\
Engineering & 1328 & 8.41 \\
Chemical Science & 1182 & 7.48 \\
Computer Science & 896 & 5.67 \\
Others & 9 & 0.02 \\
\midrule
\textbf{Question Type} & & \\
\addlinespace[0.5em]
Open-ended & 14287 & 90.38 \\
Multiple-choice & 1436 & 9.80 \\
Fill-in-the-blank & 1 & 0.01 \\
True/False & 82 & 0.52 \\
Yes/No & 2 & 0.01 \\
\midrule
\textbf{Error Type} & & \\
\addlinespace[0.5em]
Consistency Error & 5673 & 35.89 \\
Nonsensical Content & 4349 & 27.51 \\
Missing Information & 3960 & 25.05 \\
Incorrect Information & 1826 & 11.55 \\
\bottomrule
\end{tabular}
\vspace{-2mm}
\caption{Statistics of \dataname{}.}
\label{tab:datastatistics}
\end{wraptable}
\paragraph{Annotation Protocol.} After filtering, each remaining question is independently annotated by two trained annotators. Annotators judge whether the prompt constitutes a \emph{hard faulty problem}—a question that is scientifically meaningful in style but fundamentally \textbf{unsolvable or ill-posed} as written. Annotations are performed independently to reduce confirmation bias. Using the two judgments, we partition data into three groups:
\begin{itemize}
    \item \textbf{Full-false set.} Questions labeled as hard faulty by \emph{both} annotators.
    \item \textbf{Semi-false set.} Questions labeled as hard faulty by \emph{exactly one} annotator; these include borderline cases and broaden the diversity of failure modes.
    \item \textbf{Discarded set.} Questions deemed unsuitable by \emph{both} annotators (e.g., too ambiguous to adjudicate, not scientific after all, or too trivial).
\end{itemize}
Following this protocol, we obtain 2{,}610 full-false questions, 13{,}198 semi-false questions, and 2{,}183 discarded questions.

\paragraph{Data Diversity \& Statistics.}
Table~\ref{tab:datastatistics} presents dataset statistics along three dimensions: discipline, question type, and error type. \dataname{} exhibits strong diversity across scientific domains as well as a wide range of error categories.

\paragraph{Training/Dev/Test Data Split.} We construct a balanced $Test$ set by uniformly sampling 125 questions from each of 8 disciplines in the full-false subset (1,000 total), and a $Dev$ set with 50 per discipline.  The remaining data (including the semi-false subset and the remaining full-false instances) forms the training pool, resulting in 14{,}408 training problems. This split enables us not only to evaluate LLMs but also to develop and train models, in contrast to prior work, e.g., \citep{rahman2025blindsolverslogicalthinkers}, that focuses solely on evaluation benchmarks.

\section{Addressing $\mathcal{Q}_1$: Analysis of the Discriminative–Generative Discrepancy through \dataname{}}\label{sec:dataanalysis}
In this section, we analyse the performance and behavior of existing mainstream LLMs on \dataname{}, try to find some clues and insight to better understand LLMs, and also provide insight to endorse our solution design in Section \ref{sec:method}. Particularly, we want to answer the following two questions ($\mathcal{Q}_{1.1} \& \mathcal{Q}_{1.2}$):

\paragraph{$\mathcal{Q}_{1.1}$: How differently do existing LLMs behave under three prompting conditions: (1) no hint, (2) with a hint, and (3) explicit discriminative evaluation?}
 Therefore, we explore three types of prompts:
 
$\bullet$ \textbf{$P^{Gen}_{-hint}$}: we feed the question into the LLM without any hint that the problem might have issues and you should check before solving. This is the most natural LLM-asking format we want to explore and it can show the true reasoning capability of LLM. For example
        \begin{tcolorbox}[colback=gray!10, boxrule=0pt, arc=2pt]\textit{Please answer the following question: [QUESTION]}\end{tcolorbox}  
$\bullet$ \textbf{$P^{Gen}_{+hint}$}: in this case, we provide a subtle hint to above $P^{Gen}_{-hint}$: 
    \begin{tcolorbox}[colback=gray!10, boxrule=0pt, arc=2pt]\textit{Please answer the following question: [QUESTION]. If you cannot answer due to issues in the question, explicitly state what prevents a valid answer in your response}\end{tcolorbox} 
$\bullet$ \textbf{$P^{Dis}$}: Differing from above two generative prompt which are problem-solving oriented, This is a discriminative prompt to explicit ask LLM if the input question has mistake or not, such as 
        \begin{tcolorbox}[colback=gray!10, boxrule=0pt, arc=2pt]\textit{Please analyze the following question: [QUESTION]. Tell us if it has an issue (Yes or No) that prevents a valid answer and explain why.}\end{tcolorbox}  

We compare a top-performing closed-source LLM, GPT4, and three representative open-source LLMs:  Mixtral-8x7B, Llama3.3-70B and Qwen2.5-72B.

\begin{wraptable}[10]{r}{0.48\textwidth}
\vspace{-3mm}
\centering
\small
    \begin{tabular}{lccc}
        Model &  $P^{Gen}_{-hint}$ & $P^{Gen}_{+hint}$  & $P^{Dis}$\\\hline
        Mixtral-8x7B & 10.2 & 27.8 & 75.2\\
        Llama3.3 70B  & 14.0 &24.5 & 90.8\\
        Qwen2.5-72B & 10.8 & 40.1 & \textbf{95.0}\\
        GPT4 & \textbf{34.9} & \textbf{48.6} & 88.1\\\hline

    \end{tabular}
    \vspace{-2mm}
    \caption{Answer to $\mathcal{Q}_{1.1}$--performance (\%) of existing LLMs in recognizing the issues in the questions of \dataname{} under different prompt style.}
    \label{tab:resultsofthreeprompts}
\end{wraptable}
Table \ref{tab:resultsofthreeprompts} summarized the results. Here we found that i) without any hints (even though we hope any AI can recognize by default) this challenge all LLMs with GPT4 only one who passes 30\%. Adding hints in prompt can improve the potential of recognizing questions' issues with big margin ($P^{Gen}_{-hint}$ vs. $P^{Gen}_{+hint}$). ii) Interestingly, open-source models such as LLaMA and Qwen exhibit comparable or even stronger discriminative performance than GPT-4 in identifying issues in \dataname{}. This suggests that these models can serve as self-contained knowledge sources during generation, rather than relying on stronger closed-source teachers, which directly inspires our model design in Section~\ref{sec:method}. iii) both generative prompts, no matter it has hint (\textbf{$P^{Gen}_{+hint}$}) or not (\textbf{$P^{Gen}_{-hint}$}), are much behind with large margins behind the explicitly ask the LLM to discminatively check if the question has the issues (\textbf{$P^{Dis}$}). This indicates that \textbf{LLMs may has the capbility of recognizing the latent issues but it may not able to make use of it in the default autoregressive generation process}. This leads to our next question $\mathcal{Q}_{1.2}$. 

\paragraph{$\mathcal{Q}_{1.2}$: Why cannot LLMs’ reasoning
recognize the faultiness of those
questions?}  To investigate this phenomenon, we have manually compared 646 pairs of GPT4 responses under $P^{Gen}_{-hint}$ and $P^{Dis}$, where model find the question problematic based on $P^{Dis}$ but failed to reflect that when answer under $P^{Gen}_{-hint}$. Three main patterns on this symptoms are found: 

$\bullet$ \textbf{Arithmetic-over-World-Knowledge Bias (ratio: $\sim$35\%)}:   The LLMs prioritize performing surface-level arithmetic operations over validating the problem’s feasibility under basic world knowledge constraints (e.g., physical limits, inventory bounds, or domain rules). Example below:

\begin{tabular}{@{} l p{0.9\linewidth} @{}}
\toprule
\multirow{2}{*}{Q} &   Consider an amalgamation of sum number of decks of cards initially with 543 club cards. Jake has 561 club cards in their hand from the deck and then returns 111 to the deck. How many club cards are in the deck after these actions?
\\
\hdashline
$P^{Dis}$  & The question incorrectly posits a situation where the number of club cards Jake holds is more than the total available within a set, then leading to an illogical count. It is a logical contradiction as a standard deck of cards only contains 13 club cards.
 \\\hdashline
$P^{Gen}_{-hint}$  & 93
\\
\bottomrule
\end{tabular}

Many LLMs adopt a computation-first heuristic: they interpret the question as a routine numerical word problem and directly execute the arithmetic operation 543-561+111=93,
instead of validating the premise against basic world knowledge constraints (e.g., card-count limits).


$\bullet$ \textbf{Partial-Task Satisfaction (Answering the answerable component) (ratio: $\sim$25\%)}:    The question may be invalid globally (missing info, inconsistent assumptions), but it contains some valid local structure, e.g., a definable term,
a small factual claim, a solvable sub-expression, an easy “mini-problem.” Generative decoding then takes the path of least resistance: solve the “easy sub-part” and ignore the invalid piece. Instead of rejecting ill-posed problems, LLMs often “satisfice” by solving an internally identified solvable fragment and treating it as the intended task. Example below: 

\begin{tabular}{@{} l p{0.9\linewidth} @{}}
\toprule
\multirow{2}{*}{Q} &   Quenching is known to be the same as Tempering. Explain how Enthalpy applies to metallurgical processes and its importance in controlling reactions and phase transformations.
\\
\hdashline
$P^{Dis}$  & The initial statement, 'Quenching is known to be the same as Tempering,' is a factual error. Quenching and tempering are different metallurgical processes with distinct purposes, even though they are usually connected. This incorrect information makes the question misleading.
 \\\hdashline
$P^{Gen}_{-hint}$  & Enthalpy, a measure of the total energy of a thermodynamic system, is critical in metallurgical processes. It helps in determining heat exchange during reactions, which is vital in controlling phase transformations like hardening, tempering, or quenching in metallurgy. Lowering enthalpy can drive reactions towards more stable, lower-energy states, enabling metallurgists to manipulate the properties of the metal.
\\
\bottomrule
\end{tabular}

In this example, the prompt contains a false premise (“Quenching is the same as Tempering”) but also a valid sub-question about how enthalpy affects metallurgical processes. The generative model satisfices by answering the easy, well-defined part (enthalpy and phase transformations) while ignoring the invalid statement, instead of rejecting or correcting the prompt. This is partial-task satisfaction: solving a locally answerable fragment rather than addressing the full ill-posed question.

$\bullet$ \textbf{Helpfulness Preference over Non-Commitment) (ratio: $\sim$12\%)}:   Even when the model ``know'' the question is broken, it tends to avoid: ``I can’t answer'', ``the question is invalid'',  or ``insufficient informatio''. Instead, it produces something that sounds useful: a guess, a broad explanation, a plausible answer. This is not the same as instruction-following; it is a social alignment / helpfulness heuristic. This strengthens the tradeoff framing: discriminative mode licenses criticism while generative mode prioritizes helpfulness and completion. So ``knowledge exists,'' but policy preference suppresses it. LLMs exhibit refusal-avoidance in solve-mode: even when internal signals indicate ill-posedness, the model prefers producing a plausible completion to preserve helpfulness.  Example below 
\begin{tabular}{@{} l p{0.9\linewidth} @{}}
\toprule
Q & What are the uses of the folded layers of rock where geologists look for oil are called synclines?
\\
\hdashline
$P^{Dis}$  & The question is not coherent and too grammatically flawed to deduce a clear meaning. It appears to mix two independent concepts: the uses of folded layers of rock and the term 'synclines', which signifies a type of rock formation.
 \\\hdashline
\multirow{4}{*}{$P^{Gen}_{-hint}$ } & The main use of synclines in the field of geology is for the exploration and extraction of natural resources such as oil and gas. These folded layers of rock often form reservoirs where oil and gas can accumulate, making them highly important in the petroleum industry.
 \\
\bottomrule
\end{tabular}

In the above example, when the question is grammatically ill-formed or semantically ambiguous, the model refrains from explicitly stating that the prompt is confusing or underspecified. Instead, it prioritizes a “helpful” completion by generating a high-probability, generic template answer, which can introduce domain inaccuracies (e.g., conflating synclines with anticline-style structural trapping).

\section{Addressing $\mathcal{Q}_2$: Proposing a Solution--\modelname{}}
\label{sec:method}

\subsection{\modelname{} Model Construction}

\begin{figure}[t]
    \centering
    \includegraphics[width=0.9\textwidth]{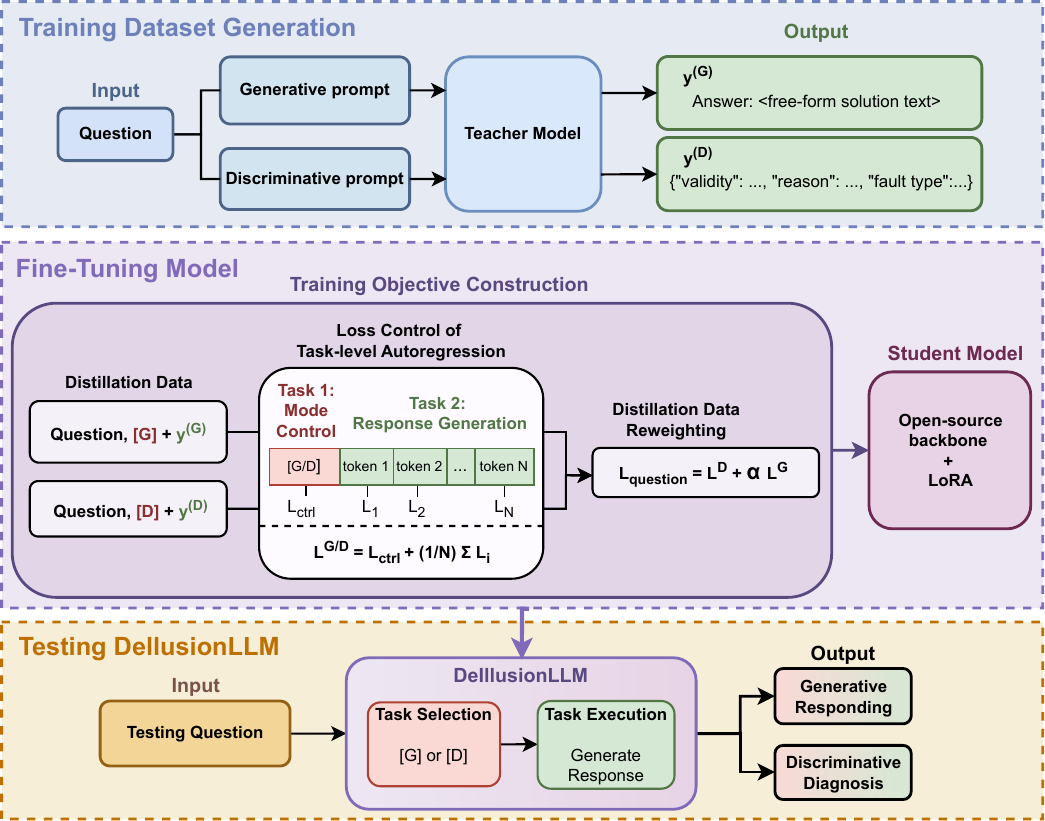}
    \vspace{-3mm}
    \caption{Overview of the proposed \modelname{} method.}
    \label{fig:overview}
    \vspace{-5mm}
\end{figure}

\paragraph{Overview.}
We propose \modelname{}, illustrated in Figure \ref{fig:overview}, a model designed to perform \emph{self-validating problem solving} via \emph{task-level autoregressive reasoning}. Instead of directly generating an answer, \modelname{} explicitly sequences two stages:
\begin{itemize}
    \item \textbf{Stage~I (Task mode selection).} Decide whether to perform \emph{discriminative diagnosis} (validate the prompt) or \emph{generative responding} (solve the problem).
    \item \textbf{Stage~II (Task execution).} Produce the corresponding output in a mode-specific format (structured validity judgment or a direct answer).
\end{itemize}
This design targets the core failure mode of standard LLMs: unconditional token-level generation that does not reliably invoke prompt validation. 
We implement task sequencing using explicit control tokens that encode the selected mode. Each training target begins with a single \emph{control token} followed by content tokens:
\begin{itemize}
    \item \textbf{Generative mode} (\texttt{[G]}): \texttt{[G]} \;\; \texttt{Answer: <free-form solution text>}
    \item \textbf{Discriminative mode} (\texttt{[D]}): \texttt{[D]} \;\; \texttt{\{"validity": ..., "reason": ...\}}
\end{itemize}
At inference time, the model first emits either \texttt{[G]} or \texttt{[D]}, then continues autoregressively to generate the mode-specific output. This yields an explicit, inspectable decision boundary between ``diagnose'' and ``answer.''

\paragraph{Distillation Data.}
We construct a supervised training corpus via distillation, using either the base model being fine-tuned (self-distillation) or a stronger external teacher model, on the \dataname{} training split. For each question $x$, we query the distillation source with a generative prompt $P^{Gen}_{-hint}$ and a discriminative prompt $P^{Dis}$, obtaining outputs $y^{(G)}$ and $y^{(D)}$, respectively. We then create two supervised training instances per question:
\[
(x,\; \texttt{[G]} \oplus y^{(G)}) \quad \text{and} \quad (x,\; \texttt{[D]} \oplus y^{(D)}),
\]
where $\oplus$ denotes concatenation. This ensures the student learns both to solve valid questions and to diagnose invalid ones, while aligning the diagnostic behavior with a high-quality ``teacher''. The resulting two-mode supervision further motivates our subsequent mode-specific and mode-composed training strategies.

\paragraph{Mode-Specific Training Objective.}
A key challenge is ensuring the model reliably learns the \emph{task-mode-selection} behavior represented by the first control token. In standard next-token training, the loss is averaged over all response tokens, which can dilute the learning signal for a single control decision, especially for long answers.

To address this, we use a token-structured objective that decouples the \textit{control-token} loss from the \textit{content-token} loss. For a target response consisting of a control token followed by $N$ content tokens, we compute:
\begin{equation}\label{eq:lossintwo}
\mathcal{L}^{mode} \;=\; \mathcal{L}_{\text{ctrl}} \;+\; \frac{1}{N}\sum_{i=1}^{N}\mathcal{L}_{i},
\end{equation}
where mode=$D$ or $G$. $\mathcal{L}_{\text{ctrl}}$ is the cross-entropy loss on the first supervised token (the control token), and $\mathcal{L}_{i}$ is the loss on the $i$-th content token. This objective prevents the control decision from being overshadowed by the rest of the output, while maintaining stable optimization for content via averaging.

\paragraph{Mode-composed Data Reweighting.} Because the diagnostic behavior is the critical missing capability under natural prompting, we optionally apply dataset-level weighting when mixing the two supervision types. Concretely, we assign higher weight to \texttt{[D]} examples than \texttt{[G]} examples during training to prioritize robust solvability judgment while still retaining generative competence. This weighting is applied at the per-example loss aggregation stage and can be tuned as a hyperparameter.
\begin{equation}\label{eq:weight}
    \mathcal{L}_{question} =  \mathcal{L}^{D}+ \alpha \mathcal{L}^{G}
\end{equation}
where $\alpha\in[0,1]$, meaning we gives default weight 1.0 to discriminative response ($D$) while decaying the loss of generative response ($G$) by $\alpha$. It is experimentally tuned on dev set.

The student model is trained on the union of \texttt{[G]} and \texttt{[D]} distillation instances constructed from the \dataname{} training split. This setting preserves the base model’s general capabilities while specializing it toward conditional problem solving.

Our approach is intentionally minimal: we do not require external verifiers, tool calls, or multi-agent scaffolding. Instead, \textit{we embed the missing step, i.e., prompt validation, into the model’s own generation process as an explicit first decision}. This directly targets the empirical gap between discriminative validity checking and generative answering, enabling the model to reliably invoke its fault-recognition knowledge before producing content.

\vspace{-3mm}
\subsection{Analyzing the effectiveness of  \modelname{}}\label{sec:modelanalysis}

\paragraph{Setup.} 
Given an input question $x$, the model should decide whether $x$ is well-posed and solvable then respond accordingly. Under the most realistic setting, we evaluate with a standard generative prompt that does \emph{not} warn the model that the question may be invalid. We fine-tune the open-source Qwen2.5-72B using LoRA \citep{DBLPHuSWALWWC22}.  A prediction is considered successful, judged by GPT-5, if the response explicitly recognizes the faultiness and refrains from producing a substantive answer that assumes corrected premises. Hyperparameter $\alpha$ in Equation \ref{eq:weight} is set to 0.6, after tuned on dev set.

For this part, we want to answer the following questions ($\mathcal{Q}_{2.1}\sim\mathcal{Q}_{2.4}$):

\vspace{-3mm}
\paragraph{$\mathcal{Q}_{2.1}$: Can \modelname{}, distilled from Qwen or GPT-4o, outperform its ``teachers'' (Qwen or GPT4) and other systems on \dataname{} under natural prompting ($P^{Gen}_{-hint}$), while preserving general reasoning ability?}\mbox{}\\

\begin{table}[t]
\centering
\begin{tabular}{lccc}
    \hline
    Model & \dataname{} ($P^{Gen}_{-hint}$) & BBH & GPQA \\
    \hline

    Mixtral-8x7B & 10.2 & 16.58 & 30.13 \\
    Llama3.3 70B & 14.0 & 17.11 & 21.88 \\\cdashline{1-4}
Qwen2.5-72B & 10.8 & 42.78 & 34.15\\ 
\modelname{} (self-distilled from Qwen) & \textbf{67.8} & 41.71 & 33.93 \\\cdashline{1-4}
    GPT-4 & 34.9 & \textbf{69.50} & 34.20 \\ 
    
\modelname{} (distilled from GPT) & 42.7 & 68.98 & \textbf{38.80} \\
    \hline
\end{tabular}
\caption{\modelname{} vs. baseline LLMs in three benchmarks.}
\label{tab:generalperformance}
\end{table}

\vspace{-3mm}
To answer this question, in addition to report on \dataname{}, we also include two regular AI reasoning benchmarks: i) \textbf{BBH} - casual judgment \citep{suzgun-etal-2023-challenging}:
    187 multiple choice questions dataset, that ask the model to judge questions about causation. We just gave the question to the model input without additional prompt. If the model were able to answer the question same as the ground truth answer, then we consider that the model got that question correct. ii) \textbf{GPQA}  \citep{rein2024gpqa}: total 448 multiple choice hard questions. We followed the zero-shot prompt from the original paper. If the model were able to answer the question same as the ground truth answer, then we consider it as correct.  

Table~\ref{tab:generalperformance} reports the results. We highlight three key findings.
(i) Incorporating discriminative capability into generative reasoning via distillation is highly effective. Both variants of \modelname{} substantially outperform their respective teachers on \dataname{} under natural prompting ($P^{Gen}_{-hint}$), with the Qwen-distilled model achieving the largest gain (67.8 vs.\ 10.8). Notably, self-distillation from Qwen is more effective than distillation from GPT-4, which is consistent with Qwen’s stronger discriminative performance observed in Table~\ref{tab:resultsofthreeprompts}.
(ii) These gains do not come at the expense of general reasoning. \modelname{} maintains performance on BBH and GPQA comparable to its teachers, with only minor variations; in particular, GPT-distilled \modelname{} even improves over GPT-4 on GPQA (38.8 vs.\ 34.2).
(iii) Overall, the results demonstrate that self-distillation is an effective strategy to mitigate the discrepancy captured by \dataname{} while preserving general reasoning ability, yielding substantial task-specific gains without degrading broader capabilities.



\begin{table*}[t]
    \centering
    \resizebox{\linewidth}{!}{%
    \begin{tabular}{c|cccccccc|c}
      Model   & Earth & Physics & Social & Math & Bio & Eng. & Chem. & CS & mean\\\hline
       GPT-4  & 37.6 & 18.4 & 56.0 & 33.6 & 31.2 & 58.4  & 24.0 & 7.2 & 33.3 \\
       \modelname{}  & 60.8 & 19.2 & 56.8 & 30.4 & 46.4 & 72.0 & 22.4 & 8.8 & 39.6\\
    \end{tabular}}
    \vspace{-3mm}
    \caption{Model performance, under $P^{Gen}_{-hint}$,  on different disciplines of \dataname{}. }
    \label{tab:cross-disciplinary-performance}
    \vspace{-3mm}
\end{table*}

We further dive into each disciplines and compare our model \modelname{} against the top competitor GPT4. Table \ref{tab:cross-disciplinary-performance} shows that GPT4-distilled \modelname{} can beat GPT4 on all eight disciplines except tiny drop on Math and Chemistry. This show our technique result in very consistent improvement across disciplines.

\vspace{-3mm}
\paragraph{$\mathcal{Q}_{2.2}$: Does task-level autoregression behave as intended?
Specifically, does the model appropriately select the diagnostic mode (\texttt{[D]}) more frequently on \dataname{} than on BBH and GPQA, and are the generated responses consistent with the selected mode (i.e., \texttt{[D]} followed by diagnostic outputs and \texttt{[G]} followed by generative answers)?}\mbox{}\\

\begin{wrapfigure}[14]{r}{0.45\textwidth}
    \centering
    \vspace{-5mm}
    \includegraphics[width=\linewidth]{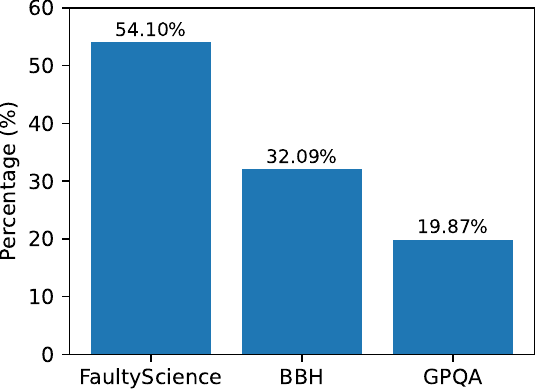}
    \vspace{-6mm}
    \caption{Mode [D] percentages for each dataset.}
    \label{fig:Q6_i_D_percentages}
    \vspace{-5mm}
\end{wrapfigure}
We first report the proportion of triggered discriminative modes (\texttt{[D]}) across datasets. Intuitively, \dataname{} should induce more discriminative behavior than BBH and GPQA, since the latter do not contain problematic questions. As shown in Figure~\ref{fig:Q6_i_D_percentages}, \dataname{} triggers \texttt{[D]} in 54.1\% of cases, broadly consistent with its overall performance (42.7\%) in Table~\ref{tab:generalperformance}. The lower task accuracy compared to the \texttt{[D]} rate suggests that some instances correctly select the diagnostic mode but still produce generative-style responses. This motivates examining the consistency between the first-stage mode selection (\texttt{[D]} vs.\ \texttt{[G]}) and the second-stage response.

Table~\ref{tab:mode_response_alignment} summarizes mode–response alignment across the three datasets. Overall, aligned cases dominate misaligned ones (72.5\% vs.\ 27.5\% on \dataname{}, 65.78\% vs.\ 34.22\% on BBH, and 80.13\% vs.\ 19.87\% on GPQA), indicating that \modelname{} generally follows consistent task-level autoregressive reasoning. Moreover, within the aligned category, \dataname{} predominantly yields diagnostic responses (\texttt{[D]} followed by a discriminative judgment), whereas BBH and GPQA mostly produce standard generative outputs (\texttt{[G]} followed by answers). Together, Figure~\ref{fig:Q6_i_D_percentages} and Table~\ref{tab:mode_response_alignment} suggest that \modelname{} largely behaves as intended.

\begin{table}[t]
\centering
\begin{tabular}{lccc}
    \toprule
    Category & \dataname{} ($P^{Gen}_{-hint}$) & BBH & GPQA \\
    \hline
    \multicolumn{4}{l}{\textit{Aligned (mode = response)}} \\
    \texttt{[D]} + response\_D & 51.70\% & 5.35\%  & 0.00\% \\
    \texttt{[G]} + response\_G & 20.80\% & 60.43\% & 80.13\% \\
    \midrule
    \multicolumn{4}{l}{\textit{Misaligned (mode $\neq$ response)}} \\
    \texttt{[D]} + response\_G & 2.40\%  & 26.74\% & 19.87\% \\
    \texttt{[G]} + response\_D & 25.10\% & 7.48\%  & 0.00\% \\   
    \toprule
\end{tabular}
\caption{Mode–response alignment distribution across datasets.}
\label{tab:mode_response_alignment}
\vspace{-7mm}
\end{table}

\vspace{-3mm}
\paragraph{$\mathcal{Q}_{2.3}$: Whether the task-level autoregression in Equation \ref{eq:lossintwo} and the distillation data reweighting in Equation \ref{eq:weight} contribute?} \mbox{}\\

\begin{wrapfigure}[10]{r}{0.45\textwidth}
    \centering
    \vspace{-7mm}
    \includegraphics[width=\linewidth]{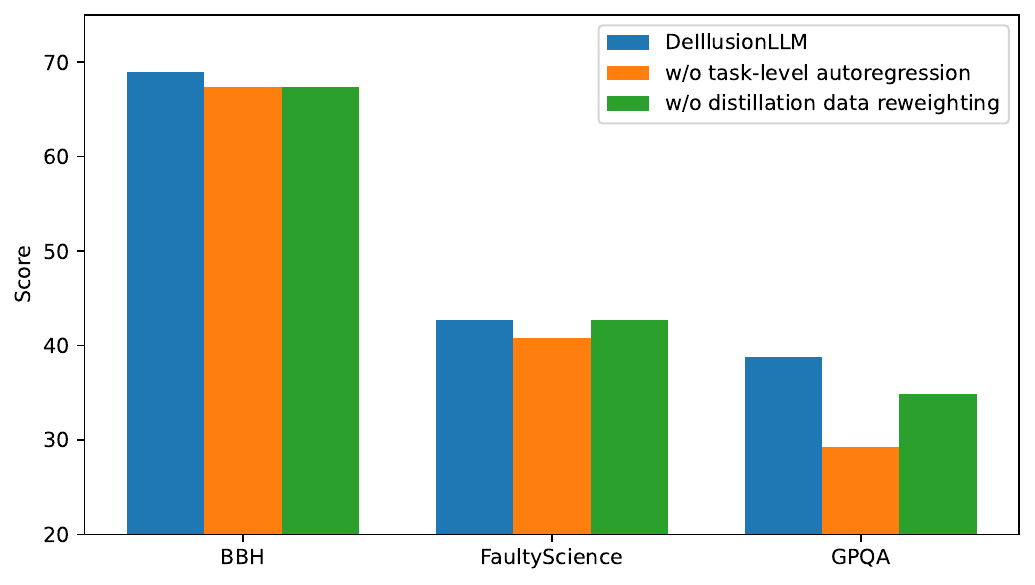}
    \vspace{-8mm}
    \caption{Ablation study.}
    \label{fig:ablation12}
    \vspace{-5mm}
\end{wrapfigure}
We conduct two ablation experiments to examine the contribution of our loss design. First, we retain the overall framework but replace the proposed loss in Equation~\ref{eq:lossintwo} with the standard LLM loss computed over the entire output sequence, including the mode token. Second, we remove the data reweighting by setting $\alpha = 1.0$ in Equation~\ref{eq:weight}, so that discriminative and generative instances receive equal weight. As shown in Figure~\ref{fig:ablation12}, both the task-level autoregressive objective and the data reweighting contribute to performance improvements, with the task-level autoregressive loss providing the largest gains. Interestingly, this effect is particularly pronounced on BBH and GPQA.

One possible explanation is that the task-level autoregressive loss strengthens the learning signal for the control token, forcing the model to make an explicit task decision before generating content. This improves the stability of mode selection and reduces ambiguity in the decoding process. For standard reasoning benchmarks such as BBH and GPQA, where almost all questions are valid and require direct solution generation, the model consistently selects the generative mode (\texttt{[G]}). As a result, the clearer task boundary introduced by the control-token loss leads to more reliable reasoning trajectories and better downstream answer generation (as shown the better performance of \modelname{} on GPQA in Table \ref{tab:generalperformance}). In contrast, on \dataname{}, performance is additionally constrained by the difficulty of detecting subtle faults, so the relative improvement from the loss design appears smaller.

\vspace{-4mm}
\paragraph{$\mathcal{Q}_{2.4}$: Any error patterns that \modelname{} still makes?} To answer this question, we analyzed all error cases in different disciplines and find the following cross-disciplinary error patterns:

$\bullet$ \textbf{Closest-answer selection under malformed multiple-choice (MCQ option-forcing)}. The model assumes one option must be right and picks/outputs something—even if the option set is inconsistent. Examples:

\begin{tabular}{@{} l p{0.88\linewidth} @{}}
\toprule
\multirow{3}{*}{Physics} & Answer the following multiple choices question.  Question: An atom has filled n = 1 and n = 2 levels. How many electrons does the atom have? Choices:  A: 2;  B: 4;  C: 6" \textbf{model\_answer}: "C: 6" (Correct Answer: 8)
\\
\hdashline
\multirow{3}{*}{Eng.}  & "Answer the following multiple choices question.  Question: If a Hexadecimal number needs to convert to binary. For each hexadecimal digit, there will be how many bits  Choices:  A: 1;  B: 2;  C: 8" \textbf{model\_response}: "4" (Correct Answer: 4)
 \\\hdashline
\multirow{6}{*}{Chem. } & "As the pH value of a neutral solution increases   (A) basic property decreases and number of OH- ions increases (B) acidic property increases and number of H+ ions decreases (C) basic property increases and number of OH- ions increases (D) acidic property decreases and number of H ions increases. Select the correct option" \textbf{model\_response}: "(C) basic property increases and number of OH- ions increases"
 \\
\bottomrule
\end{tabular}
\\

These examples illustrate the dominance of instruction-following behavior and dataset priors: because multiple-choice questions typically contain a correct option, the model tends to follow a learned ``select the best candidate'' strategy rather than validating whether the question itself is well-posed.

$\bullet$ \textbf{Silent typo-repair / anomaly suppression (especially sign errors)}. When a single token is “impossible” (negative moles/weights/ratios), the model treats it as noise and answers the repaired version. Examples

\begin{tabular}{@{} l p{0.89\linewidth} @{}}
\toprule
\multirow{3}{*}{Chem.} & "You have aquired \textcolor{red}{-2} moles Magnesium. With this in mind What is the molecular formula of C6H12O6 and how do you name it?" \textbf{model\_response}: "The molecular formula is C6H12O6 and it is named glucose." (“\textcolor{red}{-2} moles” ignored.)
\\
\hdashline
\multirow{5}{*}{Eng.}  & "A robot arm is designed to pick up two types of objects: small gears and large gears. Each small gear weighs \textcolor{red}{-2} kg, and each large gear weighs \textcolor{red}{-5} kg. The robot arm can carry a total of 12 objects at a time and can lift a total weight of \textcolor{red}{-40} kg. How many small gears and large gears can the robot arm carry?" \textbf{model\_response}: "8 small gears and 4 large gears." (model solves as if positive.)
 \\\hdashline
\multirow{3}{*}{Math } & "My grade for the midterm and Jack's grade are in the ratio \textcolor{red}{-1:1}. If Jack gets 60, what is my grade?" \textbf{model\_response}: "-60" (sometimes it accepts the negative; other times it ignores it—both indicate weak validation).
 \\
\bottomrule
\end{tabular}

These examples reflect two factors: a strong prior that user inputs may contain typos and a completion policy that favors producing something useful. As a result, single-token anomalies present little resistance during decoding and are often smoothed over rather than flagged as errors.

$\bullet$ \textbf{Unjustified assumption injection to fill missing variables}. The model invents a missing constant/definition (e.g., “standard swimming pool volume”) and proceeds. Examples

\begin{tabular}{@{} l p{0.9\linewidth} @{}}
\toprule
\multirow{5}{*}{Bio} & "A giraffe drinks 50 liters of water per day. If a giraffe lives in a zoo for 10 years, how many swimming pools could be filled with the total water it drinks?" \textbf{model\_response}: " Assuming an average swimming pool holds about 200,000 liters, a giraffe drinking 50 liters of water per day for 10 years (3650 days) would drink 182,500 liters. This amount of water could fill $\sim$ 0.9125 swimming pools."
 \\
\bottomrule
\end{tabular}
\\

In the above example, token-level decoding is optimized to continue with high-likelihood arithmetic patterns; feasibility checks require a separate ``stop-and-validate'' decision that may not reliably trigger.

\section{Conclusion}

This work highlights a structural limitation of current LLMs: despite possessing the knowledge to recognize faulty premises, models frequently fail to invoke this capability during standard generative decoding, leading to answers produced without self-verification. We frame this issue as a mismatch between token-level autoregressive generation and the inherently \emph{task-level} nature of problem solving, where validation should precede answering. To study this phenomenon, we introduce \dataname{}, a benchmark of naturally occurring faulty scientific questions, and show that existing LLMs exhibit a consistent gap between discriminative fault recognition and generative behavior. We propose \modelname{}, which introduces an explicit task decision step within the autoregressive process, enabling the model to condition answering on prior validation. The resulting improvements suggest that reliable scientific AI systems may require reasoning architectures that explicitly model \emph{task sequencing}, rather than treating all prompts as directly answerable. More broadly, this perspective points toward future LLM designs in which verification, refusal, clarification, and solution generation are integrated as coordinated reasoning stages rather than implicit side effects of next-token prediction.

\bibliography{colm2026_conference}
\bibliographystyle{colm2026_conference}


\end{document}